\documentclass[10pt,a4paper]{paper}  %

\usepackage[all]{hypcap}

\usepackage[authoryear, round]{natbib}

\usepackage{nicefrac}       %
\usepackage{multirow}
\usepackage{cleveref}
\usepackage{dsfont}
\usepackage{makecell}
\usepackage{subcaption}
\usepackage{bbm}
\usepackage{wrapfig}
\usepackage{mathtools,xparse}
\usepackage{array}
\usepackage{scalerel}
\usepackage{tablefootnote}
\usepackage{tocloft}
\usepackage{adjustbox}
\usepackage{setspace}
\usepackage[most,skins,theorems]{tcolorbox}

\newtcolorbox{promptbox}{
  colback=gray!5,
  colframe=gray!70,
  boxrule=0.8pt,
  arc=2mm,
  left=10pt,
  right=10pt,
  top=10pt,
  bottom=10pt,
  width=\linewidth,
  fontupper=\ttfamily\footnotesize,
  enhanced,
}

\newboolean{showsection}
\setboolean{showsection}{true}
\makeatletter
\@namedef{ver@everyshi.sty}{}
\makeatother

\definecolor{custom_green}{rgb}{0.0, 0.5, 0.0}
\definecolor{custom_red}{rgb}{1.0, 0.01, 0.24}
\definecolor{custom_blue}{HTML}{C9DAF7}
\definecolor{custom_purple}{HTML}{D9D1E9}
\definecolor{title_blue}{HTML}{204899}
\definecolor{cite_blue}{HTML}{044dc1}
\definecolor{cite_purple}{HTML}{7406a7}

\hypersetup{
    colorlinks = true,
    citecolor = {cite_blue},
    linkcolor = {cite_purple},
    urlcolor = {cite_purple},
}

\usepackage{listings}
\usepackage{xurl}

\newenvironment{ecompact}{
    \begin{list}{$\bullet$}{
        \setlength{\parsep}{1pt}
        \setlength{\partopsep}{1pt}
        \setlength{\topsep}{1pt}
        \setlength{\itemsep}{0.5pt}
        \setlength{\parskip}{0pt plus 1pt}
        \setlength{\leftmargin}{0.15in}
    }
}{\end{list}}

\makeatletter
\def\mathcolor#1#{\@mathcolor{#1}}
\def\@mathcolor#1#2#3{%
  \protect\leavevmode
  \begingroup
    \color#1{#2}#3%
  \endgroup
}
\makeatother

\Crefformat{equation}{#2Eq.\;(#1)#3}
\Crefformat{figure}{#2Figure #1#3}
\Crefformat{assumption}{#2Assumption #1#3}
\Crefname{assumption}{Assumption}{Assumptions}

\usepackage{crossreftools}
\makeatletter
\renewcommand\footnoterule{%
  \kern 15\p@
  \hrule \@width 2in \kern 2.6\p@
  \vspace{4pt}
}
\makeatother

\title{Collaborative Reasoning Distillation via Cross-Feedback and Coherent Curation}

\author[1]{Taehoon Kim}
\author[1]{Seunggeun Cho}
\author[1]{Dongsu Han}

\affil[1]{KAIST}

\correspondingauthor{Correspondence to: \email{\{taehoon0813, sgn.cho, dhan.ee\}@kaist.ac.kr}.}

\begin{abstract}
Reasoning capabilities are critical for advancing Large Language Models, yet current approaches either require massive computational budgets or struggle to effectively distill reasoning to smaller models. Standard distillation methods rely on outcome-based rewards, failing to distinguish between sound reasoning and lucky guesses. We propose \textbf{Collaborative Reasoning Distillation (CRD)}, a framework that enhances reasoning in compact models through three innovations: (1) interactive cross-feedback where teachers iteratively critique each other's reasoning, (2) fine-grained step-wise quality assessment capturing logical validity independent of final answers, and (3) coherence-aware step stitching that synthesizes complementary strengths. Students are trained via Reasoning Quality Optimization (RQO) with budget constraints. Our model, CRD-4B, achieves 97.3\% on MATH-500 and 70.3\% on AIME'25, surpassing baselines while using only 50K training examples, up to 12 times smaller than the datasets of comparable models.
\end{abstract}

\begin{document}

\maketitle

\section{Introduction}
\label{sec:intro}
Large Language Models (LLMs) have exhibited impressive reasoning capabilities~\citep{wei2023chainofthoughtpromptingelicitsreasoning}. However, advancing reasoning abilities requires massive computational investment. Cutting-edge models demand thousands of GPUs~\citep{jiang2024megascalescalinglargelanguage}, pushing estimated hardware costs into the \$800 million range~\citep{cottier2025risingcoststrainingfrontier}. Although training smaller models from scratch appears cost-effective, the resulting systems almost always trail far behind in reasoning quality despite significant resource outlays~\citep{fu2023specializingsmallerlanguagemodels, magister2023teachingsmalllanguagemodels}.

Knowledge Distillation (KD)~\citep{hinton2015distillingknowledgeneuralnetwork} offers a practical framework for transferring capabilities from large models to compact students. While KD has been widely applied to vision and general NLP tasks, transferring complex reasoning capabilities through distillation remains challenging~\citep{li2022explanationslargelanguagemodels, tian2024answerstransferringreasoningcapabilities}. The student's reasoning ability is inherently upper-bounded by the teacher, and any limitations or reasoning flaws in the teacher are inevitably inherited by the student, creating a ceiling effect that constrains achievable performance.

Fortunately, the growing ecosystem of diverse LLMs creates an opportunity. Different models exhibit distinct strengths across reasoning domains. Some excel at mathematical deduction, others at symbolic manipulation, logical analysis, or domain-specific reasoning. While recent work has explored leveraging multiple teachers for distillation~\citep{tian2024answerstransferringreasoningcapabilities}, existing approaches rely on simple aggregation or ensemble voting, without mechanisms to identify and combine high-quality reasoning steps across teachers.

The core challenge is that standard distillation methods derive learning signals from binary correctness or format compliance, ignoring the logical validity of intermediate steps~\citep{stanton2021doesknowledgedistillationreally}. This allows reasoning shortcuts and logical hallucinations to persist as long as they coincidentally yield correct answers. While Chain-of-Thought (CoT) prompting~\citep{wei2023chainofthoughtpromptingelicitsreasoning} provides explicit reasoning paths useful for distillation~\citep{li2022explanationslargelanguagemodels}, single-teacher CoT suffers from limited diversity, bias propagation, and inconsistent quality. Recent approaches like Group Relative Policy Optimization (GRPO)~\citep{shao2024deepseekmathpushinglimitsmathematical} have made progress in aligning model outputs. Advanced reasoning models such as DeepSeek-R1~\citep{Guo_2025} demonstrate that extended test-time computation enhances reasoning capabilities. However, longer reasoning chains substantially increase inference latency and computational cost, limiting deployment in latency-sensitive applications.

To address this limitation, we propose a complementary approach. Rather than requiring extensive test-time computation, we distill insights from teacher reasoning by assessing the logical quality of intermediate steps and synthesizing high-quality reasoning patterns across multiple teachers. This allows student models to achieve comparable reasoning quality with far fewer final training trajectories and shorter reasoning at inference.

To achieve this vision, we introduce \textbf{Collaborative Reasoning Distillation (CRD)}, a framework combining multiple teachers with complementary strengths, assessing step-level reasoning quality, and synthesizing high-quality reasoning patterns. Our key contributions are:

\begin{ecompact}
    \item \textbf{Interactive Multi-Teacher Cross-Feedback}: An iterative mechanism where diverse teachers critique and refine each other's reasoning, producing a richer candidate pool with corrected errors and complementary solution strategies.

    \item \textbf{Fine-grained Step-wise Quality Assessment}: A dual-judge system (JudgeLM and xVerify) that evaluates the logical validity of each reasoning step independent of the final answer. Unlike binary outcome-based rewards, this system assigns granular quality scores to each step, enabling the identification of high-quality reasoning patterns even within incorrect solutions.

    \item \textbf{Coherence-Aware Step-Level Stitching}: A Viterbi-style algorithm that constructs hybrid reasoning chains by selecting high-quality steps across multiple teachers, enabling solutions no single teacher could produce alone.
\end{ecompact}

Together, these three components produce a curated training dataset rich in diverse, high-quality reasoning examples with explicit step-level quality signals. We train students on this dataset using Reasoning Quality Optimization (RQO) with curriculum-based budget constraints to achieve both reasoning quality and inference efficiency.

CRD substantially improves reasoning across mathematical, scientific, and programming benchmarks. On the DeepSeek-R1-Distill-Qwen-1.5B student, CRD-1.5B achieves +7.2\% improvements on MATH-500, +6.8\% on GPQA Diamond, and +10.3\% on LiveCodeBench. On Qwen3 students, CRD-1.7B gains +2.0\% on AIME'24 and +5.4\% on AIME'25 over same-parameter baselines.
Crucially, CRD achieves these results using only 50K training examples, demonstrating that high-quality step-level curation can make reasoning distillation substantially more data-efficient.

\section{Methodology}
\label{sec:methodology}

\begin{figure*}[t]
    \centering
    \includegraphics[width=0.95\textwidth]{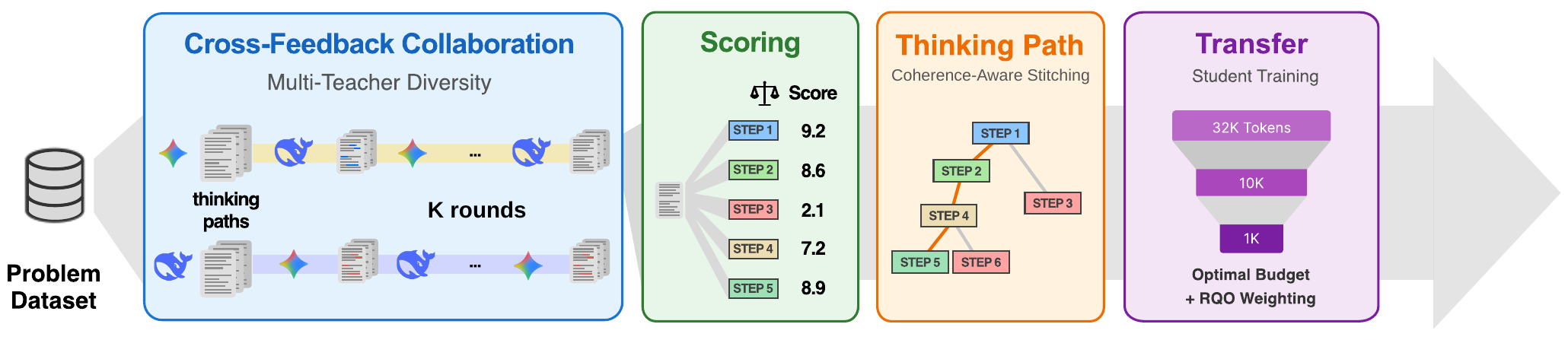}
    \caption{Overview of the CRD Framework. Multiple teachers iteratively critique and refine each other's reasoning through \textbf{Cross-Feedback}. An \textbf{LLM-as-a-Judge} then assigns step-wise quality scores, and the \textbf{Thinking Path} algorithm stitches high-quality steps across teachers while pruning flawed steps. The curated dataset is used to train students via \textbf{RQO} with Optimal Budget curriculum.
    }
    \label{fig:framework}
\end{figure*}

Current reasoning distillation methods primarily rely on imitation learning from a single teacher or outcome-based reinforcement learning (e.g., DeepSeek-R1, GRPO). While effective for final answer accuracy, these approaches treat reasoning as a monolithic sequence, failing to capture the granular quality of intermediate steps.

We introduce \textbf{Collaborative Reasoning Distillation (CRD)}, a framework that shifts from imitation to collaborative synthesis.
As illustrated in Figure~\ref{fig:framework}, CRD transforms multiple teacher outputs into a curated training dataset through three integrated components.
First, diverse teachers generate initial reasoning chains through iterative cross-feedback. Rather than treating each teacher independently, teachers critique and refine each other's outputs, producing high-quality candidates that reflect complementary strengths.
Second, we assess the logical validity of each step using a multi-judge system. Fine-grained quality scores capture reasoning quality beyond binary correctness, identifying high-quality patterns even in failed attempts.
Third, we construct hybrid reasoning chains via coherence-aware step stitching, combining high-quality steps from different teachers while maintaining logical flow.
Finally, we train students via Reasoning Quality Optimization (RQO) with curriculum-based budget constraints, enabling efficient reasoning within token limits. In our experiments, we refer to these components as Cross-Feedback, Judging, Thinking Path, and Optimal Budget.

\subsection{Interactive Multi-Teacher Cross-Feedback}
\label{sec:cross_feedback}
Recent studies reveal that LLMs exhibit complementary error profiles. Models like DeepSeek-R1 are prone to computational errors and statement skipping, omitting intermediate steps~\citep{abdollahi2025demystifyingerrorsllmreasoning}, while models like Gemini exhibit shallow reasoning with prediction bias~\citep{gao2025comparisondeepseekllms}. A student trained on a single teacher is often biased toward that teacher's error profile, motivating approaches that leverage these asymmetric weaknesses for mutual correction~\citep{du2023improvingfactualityreasoninglanguage}.

We introduce a cross-feedback loop where $N$ distinct teachers (we use $N=2$, DeepSeek-R1 and Gemini 2.5 Flash Thinking) engage in iterative critique and refinement. Rather than simple ensemble voting, each teacher reviews peer outputs, identifies logical gaps and errors, and generates improved reasoning by synthesizing complementary insights. This interactive process produces reasoning chains of higher logical depth and diversity than any single teacher could generate alone.

For the initial round ($k=0$), each teacher $T_i$ independently generates a baseline reasoning chain $R_i^{(0)}$ for the given problem $x$. In subsequent refinement rounds ($k \in \{1, 2, \ldots, K\}$), each teacher $T_i$ receives a structured input consisting of the original problem $x$ and the reasoning chains from all other teachers in the previous round $\{R_j^{(k-1)} \mid j \neq i\}$. A carefully designed review prompt instructs the teacher to examine these peer chains, identify logical gaps and errors, suggest corrections, and synthesize insights from multiple perspectives (see Appendix~\ref{app:review_prompt} for the complete prompt). The teacher then generates an improved reasoning chain $R_i^{(k)}$ that incorporates this feedback.

After $K$ rounds of cross-feedback refinement, we collect the complete set of reasoning chains including both initial and refined outputs from all teachers. Our experiments show that performance converges at $K=2$ to $3$ rounds, with additional iterations providing diminishing returns (Appendix~\ref{sec:appendix_multi_round}).

Cross-feedback substantially improves dataset quality. Comparing initial and refined outputs, mathematical reasoning accuracy increased from 64.1\% to 83.5\%, and code generation success rate improved from 52.3\% to 73.9\%. Our ablation study shows this translates to +3.4\%--7.1\% student performance improvements across benchmarks (Section~\ref{sec:ablation}).

\subsection{Step-wise Reasoning Quality Assessment}
\label{sec:quality_assessment}

Current models like DeepSeek-R1~\citep{Guo_2025} primarily rely on outcome-based rewards, where intermediate steps receive identical signals regardless of logical validity.
This sparse reward structure fails to distinguish between efficient reasoning and verbose or lucky guesses, treating all correct-answer paths equivalently.

To address this limitation, we introduce a fine-grained step-wise quality assessment system that directly evaluates the logical validity of each intermediate step, independent of the final answer. Instead of deriving rewards from format compliance or outcome correctness, we assign continuous scalar scores to every step, capturing logical validity and reasoning efficiency~\citep{lightman2023letsverifystepstep, li2025processrewardmodelqvalue}.
This enables students to learn the distinction between high-quality reasoning paths and low-quality paths that coincidentally reach the correct answer, and to internalize the nuanced differences among multiple correct solutions.

We employ two specialized judge models to evaluate each step: \textbf{JudgeLM}~\citep{zhu2025judgelmfinetunedlargelanguage}, which achieves 16× speedup over GPT-4o with comparable accuracy, and \textbf{xVerify}~\citep{chen2025xverifyefficientanswerverifier}, a lightweight 0.5B model with exceptional efficiency.

To capture algorithmic uncertainty, each judge evaluates every step twice, yielding a set of four independent scores.
To ensure robustness against stochastic outliers, we apply a \textbf{25\% trimmed mean} aggregation strategy: the minimum and maximum scores are discarded, and the remaining two intermediate scores are averaged to produce the final step quality score $v_t$.
In our ablations, using both judges provides larger gains than using either judge alone, suggesting that JudgeLM and xVerify contribute complementary quality signals rather than redundant scores (Appendix~\ref{app:judge}).

After step-wise scoring, we perform domain-specific verification combining LLM  evaluation with specialized tools.
For code, we execute solutions and penalize steps after failures. For math, we compare against ground truth and penalize steps after divergence points. For logic, we verify inference rule correctness.

\subsection{Coherence-Aware Step Stitching}
\label{sec:thinking_path}

The third component of CRD is \textbf{Coherence-Aware Step Stitching}, which we call the \textit{Thinking Path} algorithm. Standard distillation selects the best \textit{complete} chain from a single teacher, inheriting that teacher's limitations even when other teachers excel at specific steps. In contrast, CRD enables step-level synthesis: combining Teacher 1's novel problem setup with Teacher 2's rigorous derivation with Teacher 1's verification, even if neither teacher alone produces a complete correct solution.

This capability is possible because we have both fine-grained quality scores (Section~\ref{sec:quality_assessment}) and multiple diverse reasoning chains (Section~\ref{sec:cross_feedback}). The challenge is selecting steps that maintain logical coherence. Simply picking the highest-scoring step at each position creates Frankenstein chains where steps do not flow naturally into each other.

We formulate this as a coherence-aware path-finding problem on a directed acyclic graph (DAG)~\citep{yao2023treethoughtsdeliberateproblem}, where nodes represent reasoning steps $s_t^{(i)}$ from teacher $T_i$ at position $t$. We define the optimal hybrid path $\hat{R}$ as the sequence that maximizes both step-wise quality and semantic coherence between consecutive steps.

To handle granularity differences across teachers, we align steps using embedding similarity with lookahead merging: if consecutive steps from one teacher better match a single step from another, they are merged into one node. When similarity falls below a threshold, steps remain as parallel branches rather than forced alignments.

Let $S_t = \{s_t^{(1)}, \ldots, s_t^{(N)}\}$ be the set of candidate steps at position $t$. The score for transitioning from step $u \in S_{t-1}$ to step $v \in S_t$ balances two objectives:
\begin{equation}
\text{Score}(u \to v)
= \lambda \cdot \underbrace{\mathcal{V}(v)}_{\text{Quality}}
+ (1-\lambda)\cdot \underbrace{\log(\epsilon + C(u,v))}_{\text{Coherence}}
\end{equation}
where $C(u,v)$ is the transition-coherence score between consecutive steps, computed from the cosine similarity of their embeddings and rescaled to $[0,1]$, $\mathcal{V}(v)$ is the normalized step-quality score, and $\epsilon$ is a small constant for numerical stability. We use $\lambda=0.7$ based on the sensitivity analysis in Figure~\ref{fig:lambda_sensitivity}.

We find the globally optimal path using a Viterbi-style dynamic programming
algorithm:
\begin{equation}
    \hat{R} = \operatorname*{arg\,max}_{s_1, \ldots, s_L} \sum_{t=1}^{L} \text{Score}(s_{t-1} \to s_t)
\end{equation}
This enables \textit{emergent solutions} that transcend individual teacher limitations. A problem might require Teacher 1's creative insight for setup, Teacher 2's careful arithmetic for calculation, and Teacher 1's verification logic for confirmation. By combining these complementary strengths at the step level, the resulting trajectory represents the collective best reasoning that no single teacher could produce in isolation. The complete algorithm with pseudocode is provided in Appendix~\ref{app:thinking_path_algorithm}. Among 2,694 tasks with incorrect pre-stitch trajectories, 528 (19.6\%) become correct after stitching (Appendix~\ref{app:stitching}).

\subsection{Reasoning Transfer via RQO}
\label{sec:optimization}

Once the reasoning trajectories are synthesized via step-level stitching
and assigned fine-grained quality scores, we transfer these capabilities to the student model $\pi_\theta$ using a reinforcement learning approach that directly leverages the step-wise quality signals.

\paragraph{Reasoning Quality Optimization (RQO).}
Standard preference-based methods like DPO treat all preference pairs equally, ignoring the magnitude of quality differences. In contrast, RQO directly utilizes the fine-grained quality scores from Section~\ref{sec:quality_assessment} to weight learning signals. For each reasoning chain, we aggregate the step-level quality scores to obtain a trajectory-level quality signal $v_R = \sum_{t} v_t$, where $v_t$ is the quality score of step $t$. For chain pairs $(R_{win}, v_{win})$ and $(R_{lose}, v_{lose})$ where $v_{win} > v_{lose}$, we compute the quality margin $\Delta v = v_{win} - v_{lose}$ and apply exponential weighting:
\begin{equation}
    w(\Delta v) = \frac{e^{\beta \Delta v} - 1}{Z}
\end{equation}
where $\beta$ controls the sensitivity to quality differences and $Z$ is a normalization constant. The loss function becomes:
{\small
\begin{equation}
    \mathcal{L}_{\text{RQO}} = - \mathbb{E} \left[ w(\Delta v) \cdot \log \sigma \left( \pi_\theta(R_{win}|q) - \pi_\theta(R_{lose}|q) \right) \right]
\end{equation}
}
This formulation ensures that chains with larger quality margins receive stronger learning signals, enabling the student to learn not just from correct vs. incorrect distinctions, but from the nuanced quality differences captured by step-level assessment.

\paragraph{Dynamic Length Regularization.}
Training with strict token constraints from the start fails due to the learnability gap~\citep{ding2025micotabridginglearnabilitygap}: students cannot learn complex reasoning patterns when forced to be concise prematurely. Following the Long-to-Short paradigm~\citep{wu2025unlockingefficientlongtoshortllm, hammoud2025trainlongthinkshort}, we implement a curriculum-based length penalty that progressively constrains token budgets (32K$\rightarrow$10K$\rightarrow$1K). The effective reward includes an exponential penalty for exceeding the current budget $B$:
\begin{equation}
    R_{\text{budget}}(l) = R_{\text{base}} \cdot \exp(-\alpha \max(0, l - B))
\end{equation}
where $l$ is the response length and $\alpha$ controls penalty severity. This soft constraint~\citep{li2025leashadaptivelengthpenalty} enables the model to first learn correct reasoning at longer lengths, then gradually compress into efficient solutions without performance degradation.

\section{Evaluation}
\label{sec:experiments}

\textbf{Training Data.}
Our training dataset comprises 50K reasoning instances from four diverse sources:
DeepScaleR~\citep{deepscaler2025}  for mathematical reasoning, Eurus-2-RL~\citep{cui2025processreinforcementimplicitrewards} for code generation,  Reasoning Gym~\citep{stojanovski2025reasoninggymreasoningenvironments} for logical reasoning, and SCP-116K~\citep{lu2025scp116khighqualityproblemsolutiondataset}  for STEM problems (Appendix~\ref{app:dataset}).

Recent findings show that reasoning distillation exhibits a performance valley  between 1K--30K examples~\citep{he2025valleycodereasoningscaling}. Our 50K dataset size avoids this valley while remaining significantly smaller than prior  methods (220K--800K examples), enabled by multi-stage quality filtering that  ensures high-quality reasoning across the dataset.

\paragraph{Contamination Analysis.} We systematically verified training-evaluation separation. For DeepScaleR, we identified 11 overlapping problems with MATH-500 (2.2\%), which were excluded during our 50\% subsampling; AIME'24 and AIME'25 showed zero overlap.
For Eurus-2-RL (455K math, 25K code problems), we sampled 15K code problems for training and found no contamination across all code benchmarks.
LiveCodeBench v5 problems (August 2024--February 2025) post-date all code training sources: APPS (2021), CodeContests (2022), and TACO (2023).

\textbf{Benchmarks.} We evaluate on mathematics (MATH-500~\citep{hendrycks2021measuringmathematicalproblemsolving},
AIME'24~\citep{aime2024}, AIME'25~\citep{aime2025}), code (LiveCodeBench~\citep{jain2024livecodebenchholisticcontaminationfree}),
and graduate-level STEM reasoning (GPQA-Diamond~\citep{rein2023gpqagraduatelevelgoogleproofqa}). Detailed descriptions
are in Appendix~\ref{app:benchmarks}.

\textbf{Models.} We use DeepSeek-R1 and Gemini 2.5 Flash Thinking as teachers,
running $K=2$ cross-feedback rounds to generate refined reasoning chains. For
students, we evaluate DeepSeek-R1-Distill-Qwen (1.5B) and Qwen3 (1.7B, 4B, 30B-A3B MoE)
to demonstrate generalization across dense and mixture-of-experts architectures.

\subsection{Performance Across Model Sizes}
\label{sec:main_results}

A key question in reasoning distillation is whether small models can achieve
strong performance through higher-quality supervision rather than relying solely on parameter scaling. Table~\ref{tab:scale_comparison} demonstrates that CRD answers this affirmatively through step-wise quality assessment and coherence-aware synthesis, achieving competitive performance across all model sizes.

\begin{table*}[t]
\caption{Performance comparison across benchmarks (Pass@1). We report accuracy (\%) on mathematical, coding, and general reasoning benchmarks. The best results are highlighted in \textbf{bold}. \textcolor{gray}{Gray rows} indicate DeepSeek-R1 and distilled models included as reference points for parameter efficiency comparison.}
\centering
\small
\setlength{\tabcolsep}{6pt}
\begin{tabular}{@{}l c c c c c@{}}
\toprule
& \multicolumn{3}{c}{\textbf{Math}} & \textbf{Code} & \textbf{Graduate-Level STEM} \\
\cmidrule(lr){2-4} \cmidrule(lr){5-5} \cmidrule(lr){6-6}
\textbf{Model} & \textbf{MATH-500} & \textbf{AIME'24} & \textbf{AIME'25} & \textbf{LCB} & \textbf{GPQA-Diamond} \\
\midrule
\multicolumn{6}{l}{\cellcolor{gray!10}\textit{\textbf{Tiny Scale ($<$2B)}}} \\
\textcolor{gray}{DeepSeek-R1-Distill-7B} & \textcolor{gray}{92.8} & \textcolor{gray}{55.5} &\textcolor{gray}{--} & \textcolor{gray}{37.6} & \textcolor{gray}{49.1} \\
Gemini-Flash-Distill-1.5B & 85.3 & 32.5 & 24.6 & 17.2 & 31.6 \\
DeepSeek-R1-Distill-1.5B & 83.9 & 28.9 & 22.0 & 16.9 & 32.9 \\
ProRL-1.5B\textsuperscript{$\dagger$} & \textbf{91.9} & \textbf{48.1} & 33.3 & 23.8 & \textbf{41.7} \\
\textbf{CRD-1.5B\textsuperscript{$\dagger$}} & 91.1 & 46.9 & \textbf{35.1} & \textbf{27.2} & 39.7 \\
\midrule
\textcolor{gray}{DeepSeek-R1-Distill-7B} & \textcolor{gray}{92.8} & \textcolor{gray}{55.5} &\textcolor{gray}{--} & \textcolor{gray}{37.6} & \textcolor{gray}{49.1} \\
Qwen3-1.7B\textsuperscript{$\dagger$} & 93.4 & 48.3 & 36.8 & 33.2 & 40.1 \\
\textbf{CRD-1.7B\textsuperscript{$\dagger$}} & \textbf{94.5} & \textbf{50.3} & \textbf{42.2} & \textbf{37.3} & \textbf{42.9} \\

\midrule
\multicolumn{6}{l}{\cellcolor{gray!10}\textit{\textbf{Small Scale ($\leq$4B)}}} \\
\textcolor{gray}{DeepSeek-R1-Distill-32B} & \textcolor{gray}{94.3} & \textcolor{gray}{72.6} &\textcolor{gray}{60.0} & \textcolor{gray}{57.2} & \textcolor{gray}{62.1} \\
Qwen3-4B\textsuperscript{$\dagger$} & 97.0 & 73.8 & 65.6 & 54.2 & 55.9 \\
\textbf{CRD-4B\textsuperscript{$\dagger$}} & \textbf{97.3} & \textbf{76.2} & \textbf{70.3} & \textbf{57.4} & \textbf{58.7} \\

\midrule
\multicolumn{6}{l}{\cellcolor{gray!10}\textit{\textbf{Large Scale ($\geq$30B)}}} \\
\textcolor{gray}{EXAONE Deep 32B} & \textcolor{gray}{95.7} & \textcolor{gray}{72.1} & \textcolor{gray}{65.8} & \textcolor{gray}{59.5} & \textcolor{gray}{66.1} \\

\textcolor{gray}{Hermes 4 70B} & \textcolor{gray}{95.5} & \textcolor{gray}{73.5} & \textcolor{gray}{67.5} & \textcolor{gray}{50.5} & \textcolor{gray}{66.1} \\
\textcolor{gray}{DeepSeek-R1 (671B, A37B)} & \textcolor{gray}{97.3} & \textcolor{gray}{79.8} & \textcolor{gray}{70.0} & \textcolor{gray}{63.5} & \textcolor{gray}{71.5} \\
Qwen3-30B-A3B\textsuperscript{$\dagger$} & 98.0 & 80.4 & 70.9 & 62.6 & 65.8 \\
\textbf{CRD-30B-A3B\textsuperscript{$\dagger$}} & \textbf{98.3} & \textbf{82.5} & \textbf{73.2} & \textbf{65.9} & \textbf{67.2} \\
\bottomrule
\end{tabular}
\begin{flushleft}
\textsuperscript{$\dagger$} Models marked with $\dagger$ share the same baseline model; our method generally outperforms them across benchmarks.
\end{flushleft}
\label{tab:scale_comparison}
\end{table*}

At the tiny scale (1.5B), CRD-1.5B achieves 91.1\% on MATH-500 compared to the baseline DeepSeek-R1-Distill's 83.9\%. On AIME'25, CRD-1.5B reaches 35.1\%, slightly exceeding ProRL-1.5B~\citep{liu2025prorlprolongedreinforcementlearning} (33.3\%), the current state-of-the-art at the 1.5B scale, while using 2.7$\times$ less training dataset with a similar RL-based approach.
The gray rows contextualize this result: CRD-1.5B approaches the 7B distilled model on MATH-500 (91.1 vs.\ 92.8, \textbf{4.7$\times$ smaller}), while CRD-1.7B (94.5\%) surpasses the 14B distilled model (93.9\%) on MATH-500 (\textbf{8.2$\times$ smaller}). This suggests that higher-quality supervision can partly substitute for parameter scaling at small model sizes.

\begin{figure}[h]
  \centering

  \begin{subfigure}[b]{0.48\textwidth}
    \centering
    \includegraphics[width=\linewidth]{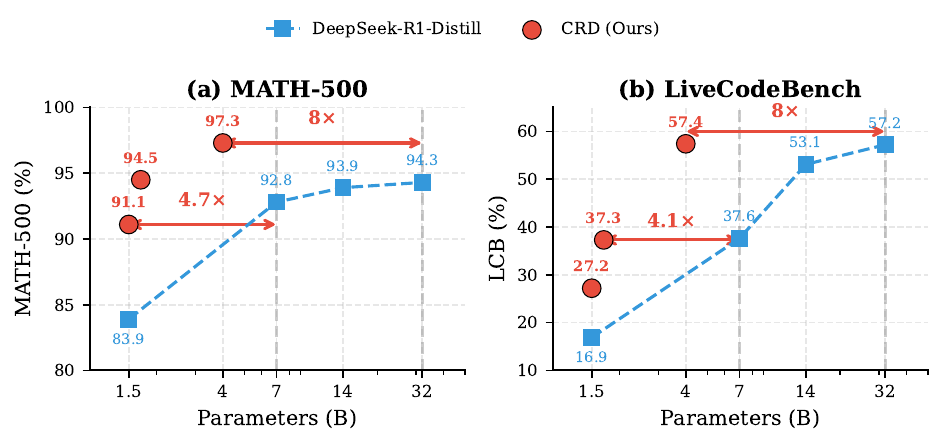}
    \caption{Parameter efficiency comparison.}
    \label{fig:scaling_loglog}
  \end{subfigure}
  \hfill
  \begin{subfigure}[b]{0.48\textwidth}
    \centering
    \includegraphics[width=\linewidth]{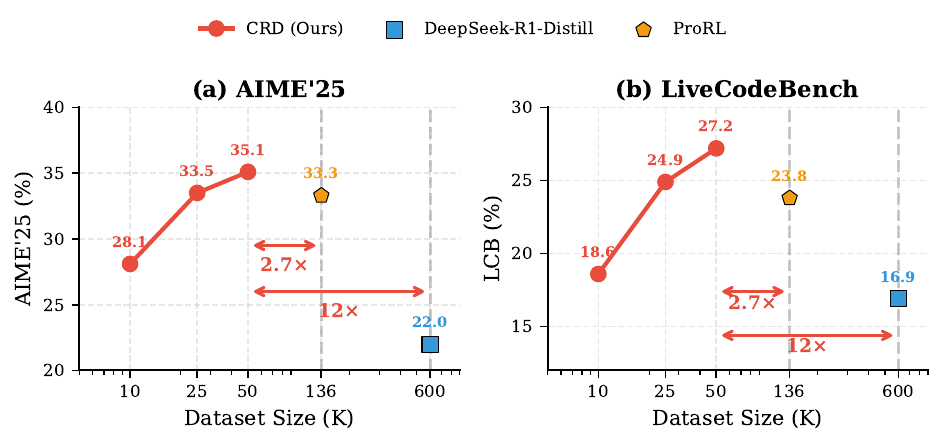}
    \caption{Data efficiency comparison.}
    \label{fig:data_efficiency}
  \end{subfigure}
  \caption{(\textbf{a}) CRD achieves comparable performance to DeepSeek-R1-Distill models with 4.7--8$\times$ fewer parameters. CRD-1.5B approaches the 7B model on MATH-500 (4.7$\times$), while CRD-4B surpasses the 32B model (8$\times$). (\textbf{b}) CRD achieves superior performance with 12$\times$ less data than DeepSeek-R1-Distill (600K reasoning samples) and 2.7$\times$ less than ProRL (136K), while outperforming both on AIME'25 and LCB.}
  \label{fig:efficiency_comparison}
\end{figure}

At the small scale (4B), CRD-4B achieves 97.3\% on MATH-500 and 70.3\% on AIME'25, approaching saturation on standard benchmarks. CRD-4B exceeds Hermes 4 70B~\citep{teknium2025hermes4technicalreport} (\textbf{17$\times$ smaller}) and EXAONE Deep 32B~\citep{bae2026exaonedeepreasoningenhanced} on AIME'25 and outperforms 32B distilled model (\textbf{8$\times$ smaller}) on all math and code benchmarks, demonstrating that CRD provides an alternative to parameter scaling. Figure~\ref{fig:scaling_loglog} visualizes this parameter efficiency across model scales.

At the large scale (30B), CRD-30B-A3B improves across diverse reasoning domains.
It improves mathematical reasoning on AIME'25 from 70.9\% to 73.2\% (+2.3\%), code generation on LiveCodeBench from 62.6\% to 65.9\% (+3.3\%), and STEM reasoning on GPQA-Diamond from 65.8\% to 67.2\% (+1.4\%).
These results show that CRD consistently improves performance across all model scales and problem types.
Remarkably, CRD-30B-A3B surpasses its \textbf{own teacher} DeepSeek-R1 on math and code benchmarks, despite using \textbf{12$\times$ fewer active parameters} (3B vs 37B).
This demonstrates that CRD enables students to exceed teacher capabilities through cross-teacher synthesis.

\begin{table}[t]
  \centering

  \begin{minipage}[t]{0.40\textwidth}
    \centering
    \caption{Multi-teacher comparison on Qwen-2.5-1.5B, Llama-3.1-8B.\\}
    \label{tab:multi_teacher_qwen}
    \scriptsize
    \setlength{\tabcolsep}{1.5pt}
    \begin{tabular}{@{}l c c c c@{}}
    \toprule
    \textbf{Method} & \textbf{GSM8K} & \textbf{MATH-500} & \textbf{HumanEval} & \textbf{LCB} \\
    \midrule
    Qwen-2.5-1.5B (Base) & 73.2 & 55.3 & 61.6 & 14.8 \\
    TinyLLM & 81.8 & 68.1 & 64.3 & 15.7 \\
    FAIR & 82.3 & 73.7 & 65.8 & 16.3 \\
    MAGDi & 85.9 & 72.6 & 64.9 & 18.4 \\
    \midrule
    \textbf{CRD (Ours)} & \textbf{95.2} & \textbf{91.1} & \textbf{72.3} & \textbf{27.2} \\
    \midrule
    Llama-3.1-8B (Base) & 84.5 & 51.9 & 72.6 & 16.2 \\
    FAIR & 86.2 & 54.8 & 75.4 & 18.5 \\
    MAGDi & 86.9 & 56.1 & 76.2 & 18.8 \\
    \midrule
    \textbf{CRD (Ours)} & \textbf{92.6} & \textbf{62.3} & \textbf{85.1} & \textbf{22.4} \\
    \bottomrule
    \end{tabular}
  \end{minipage}
  \hfill
  \begin{minipage}[t]{0.50\textwidth}
    \centering
    \caption{Ablation study on CRD components. \\Base model is DeepSeek-R1-Distill-Qwen-1.5B.\\}
    \label{tab:ablation}
    \scriptsize
    \setlength{\tabcolsep}{1.5pt}
    \renewcommand{\arraystretch}{1.52}
    \begin{tabular}{@{}lcccc@{}}
    \toprule
    \textbf{Configuration} & \textbf{MATH-500} & \textbf{AIME'25} & \textbf{LCB} & \textbf{GPQA-Diamond} \\
    \midrule
    Base Model & 83.9 & 22.0 & 16.9 & 32.9 \\
    \midrule
    + Cross-Feedback & 87.3 {\tiny\textcolor{blue}{+3.4}} & 29.1 {\tiny\textcolor{blue}{+7.1}} & 22.8 {\tiny\textcolor{blue} {+5.9}} & 36.7 {\tiny\textcolor{blue}{+3.8}} \\
    + Thinking Path & 89.4 {\tiny\textcolor{blue}{+2.1}} & 32.6 {\tiny\textcolor{blue}{+3.5}} & 25.4 {\tiny\textcolor{blue}{+2.6}} & 37.8 {\tiny\textcolor{blue}{+1.1}} \\
    + Optimal Budget & 89.7 {\tiny\textcolor{blue}{+0.3}} & 33.2 {\tiny\textcolor{blue}{+0.6}} & 25.8 {\tiny\textcolor{blue}{+0.4}} & 38.0 {\tiny\textcolor{blue}{+0.2}} \\
    + Judging \textbf{(CRD)} & \textbf{91.1} {\tiny\textcolor{blue}{+1.4}} & \textbf{35.1} {\tiny\textcolor{blue}{+1.9}} &\textbf{27.2} {\tiny\textcolor{blue}{+1.4}} & \textbf{39.7} {\tiny\textcolor{blue}{+1.7}} \\
    \midrule
    \textbf{Total Gain} & \textbf{$\uparrow$7.2} & \textbf{$\uparrow$13.1} & \textbf{$\uparrow$10.3} & \textbf{$\uparrow$6.8} \\
    \bottomrule
    \end{tabular}
  \end{minipage}
\end{table}

\paragraph{Comparison with Multi-Teacher Approaches.}
\label{sec:multi_teacher_comparison}

We compare CRD against existing multi-teacher distillation methods to isolate the benefit of step-level quality curation.
Table~\ref{tab:multi_teacher_qwen} reports results using the same Qwen2.5-1.5B and Llama-3.1-8B student backbones and the same pool of CRD-generated teacher candidates, isolating the effect of the distillation and curation strategy.
\begin{ecompact}
    \item \textbf{TinyLLM}~\citep{tian2024answerstransferringreasoningcapabilities} uses teacher-forcing CoT to transfer reasoning from multi-teachers, but does not perform step-level quality scoring.
    \item \textbf{FAIR}~\citep{li2025learningcommitteereasoningdistillation} employs peer-review between teachers to identify student mistakes and filter rationales via acceptance threshold.
    \item \textbf{MAGDi}~\citep{chen2024magdistructureddistillationmultiagent} represents multi-agent interactions as graphs and uses contrastive learning between correct and incorrect reasoning.
\end{ecompact}

CRD outperforms all baselines by substantial margins: +18.5\% over MAGDi on MATH-500, +9.3\% on GSM8K, and +8.8\% on LiveCodeBench. This gap stems from CRD's \textbf{step-level} granularity: while TinyLLM, FAIR, and MAGDi operate at the chain level, selecting or weighting complete solutions, CRD evaluates each reasoning step independently. A chain with one flawed step is entirely discarded by chain-level methods, but CRD extracts its valid steps and combines them with correct steps from other teachers. This synthesis is particularly impactful on multi-step reasoning (MATH-500: +18.5\%), where a single error propagates to incorrect answers.

\paragraph{Generalization Beyond the Training Distribution.} Table~\ref{tab:generalization} reports two held-out evaluations. On BoxNet, an out-of-distribution planning task used by ProRL and excluded from our training data, CRD-1.5B is comparable to ProRL at pass@1 and higher at pass@2 and pass@4. On the text-only subset of HLE~\citep{phan2026benchmark}, CRD-30B-A3B improves over its Qwen3-30B-A3B base by 3.23 points.

\begin{table}[h]
\centering
\caption{Generalization beyond the training distribution. Left: BoxNet planning (100 procedurally generated instances, our re-evaluation of all models, temperature 0.6, top-$p$ 0.95, 32K tokens). Right: HLE text-only subset (official protocol).}
\label{tab:generalization}
\small
\begin{minipage}[t]{0.58\textwidth}\centering
\begin{tabular}{@{}lccc@{}}
\toprule
\textbf{BoxNet} & \textbf{pass@1} & \textbf{pass@2} & \textbf{pass@4} \\
\midrule
DeepSeek-R1-Distill-1.5B & 0.0 & 0.0 & 0.0 \\
ProRL-1.5B & \textbf{7.1} & 13.8 & 22.4 \\
CRD-1.5B & 6.8 & \textbf{14.6} & \textbf{25.3} \\
\bottomrule
\end{tabular}
\end{minipage}\hfill
\begin{minipage}[t]{0.38\textwidth}\centering
\begin{tabular}{@{}lc@{}}
\toprule
\textbf{HLE} & \textbf{Text-only} \\
\midrule
Qwen3-30B-A3B & 11.31 \\
CRD-30B-A3B & \textbf{14.54} \\
\bottomrule
\end{tabular}
\end{minipage}
\end{table}

\subsection{Data Efficiency: Quality Over Quantity}
\label{sec:data_efficiency}

Figure~\ref{fig:data_efficiency} demonstrates CRD's data efficiency. At 50K samples, CRD-1.5B reaches 35.1\% on AIME'25, surpassing DeepSeek-R1-Distill (22.0\% on 600K reasoning samples) and ProRL (33.3\% on 136K samples). CRD achieves this superior performance using \textbf{12$\times$} and \textbf{2.7$\times$} less data, respectively.
Notably, CRD-1.5B exhibits consistent data scaling behavior within our curated dataset, improving monotonically from 28.1\% at 10K to 35.1\% at 50K. This curve is shifted upward by quality-aware curation, so CRD achieves with 10K samples what baselines require 600K to match.
DeepSeek-R1-Distill uses 800K total samples (600K reasoning, 200K non-reasoning); we compare against the reasoning portion for fair evaluation.

This efficiency stems mainly from curation, since naive SFT on the same 50K trajectories already reaches 90.5\% on MATH-500 (Table~\ref{tab:sft_vs_rqo}). RQO adds a consistent further gain by weighting preference pairs with their step-derived quality margins.

CRD improves reasoning distillation efficiency by replacing scale-driven supervision with quality-aware data selection. Instead of relying on a large pool of distillation corpus, CRD constructs a compact 50K-example training set using the same teacher models, yielding a 12$\times$ smaller final trajectory set and a 2.7$\times$ smaller dataset than ProRL's 136K examples. Although CRD generates multiple candidate chains per prompt through cross-feedback, only high-quality trajectories are retained for student learning.
As a result, the 1.5B student requires only 1,632 training steps and 8.8 H100 GPU-hours, compared with 7,800 steps and 42 H100 GPU-hours for the 600K distillation baseline and more than 2K steps for ProRL.
This supports our main claim that improving the quality and coherence of reasoning supervision can reduce the amount of final training data required by the student.

\subsection{Ablation Study: Understanding Each Component}
\label{sec:ablation}

Table~\ref{tab:ablation} isolates the contribution of each CRD component, revealing that they address distinct limitations in reasoning distillation. At the step level, algebra step correctness improves from 85.8\% to 92.2\% (Appendix~\ref{app:detailed_analysis}), and cross-feedback converges after two rounds (Appendix~\ref{sec:appendix_multi_round}). Appendix~\ref{app:student_outputs} shows representative student outputs.

\textbf{Cross-Feedback} provides the largest individual gain (+3.4--7.1\%), with particularly strong improvements on complex benchmarks: +7.1\% on AIME'25 and +5.9\% on LiveCodeBench. 
This aligns with our motivation that teachers exhibit complementary error profiles. DeepSeek-R1 excels at creative problem formulation but makes computational errors. Gemini provides rigorous derivations but occasionally shallow reasoning.
Cross-feedback enables mutual correction, producing refined candidates that repair errors found in individual teacher outputs.

\textbf{Thinking Path} contributes +1.1--3.5\% by enabling step-level fusion across teachers.
When Teacher 1's novel approach fails due to a logic bug and Teacher 2's standard approach fails due to a calculation error, Thinking Path stitches Teacher 1's creative setup with Teacher 2's correct computation, yielding a correct solution that neither teacher produced alone.

\textbf{Optimal Budget} and \textbf{Judging} provide incremental refinements (+0.2--0.6\% and +1.4--1.9\% respectively). Optimal Budget implements curriculum-based length constraints that help students learn efficient reasoning without premature truncation~\citep{ding2025micotabridginglearnabilitygap}. Judging adds fine-grained quality signals that enable the student to distinguish between high-quality and mediocre reasoning paths, even when both reach correct answers.

\textbf{Same-data SFT vs.\ RQO.} Table~\ref{tab:sft_vs_rqo} compares RQO with naive SFT trained on the same 50K Thinking Path trajectories from the same DeepSeek-R1-Distill-Qwen-1.5B checkpoint. Naive SFT already reaches 90.5\% on MATH-500, so most of the gain comes from the curated data. RQO adds consistent gains on all four benchmarks (+1.15 points on average).

\begin{table}[h]
\centering
\caption{Same-data comparison of training objectives (DeepSeek-R1-Distill-Qwen-1.5B, 50K Thinking Path trajectories).}
\label{tab:sft_vs_rqo}
\small
\begin{tabular}{@{}lcccc@{}}
\toprule
\textbf{Method} & \textbf{MATH-500} & \textbf{AIME'25} & \textbf{LCB} & \textbf{GPQA-Diamond} \\
\midrule
Naive SFT & 90.5 & 33.4 & 25.7 & 38.9 \\
RQO (CRD) & \textbf{91.1} & \textbf{35.1} & \textbf{27.2} & \textbf{39.7} \\
\bottomrule
\end{tabular}
\end{table}

\begin{figure}[h]
    \centering
    \begin{minipage}{0.41\textwidth}
        \centering
        \includegraphics[width=\linewidth]{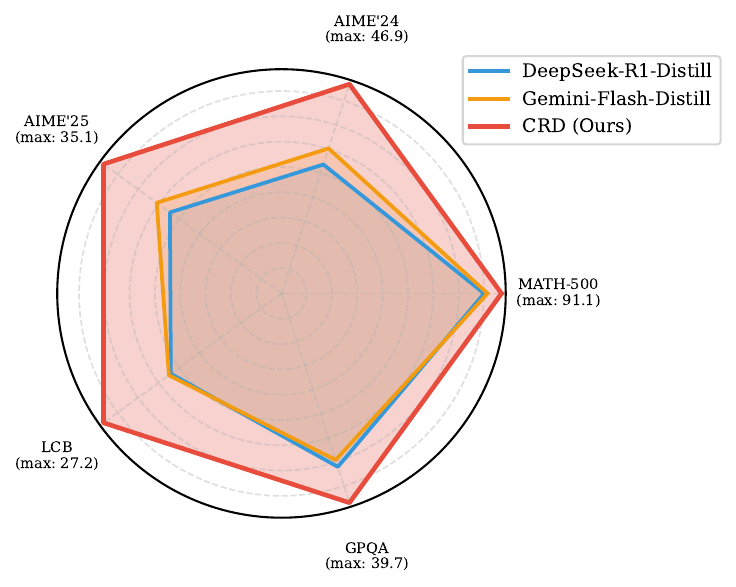}
        \caption{Single-teacher vs. multi-teacher comparison on Qwen2.5-1.5B models.}
        \label{fig:radar_single_multi}
    \end{minipage}\hfill
    \begin{minipage}{0.55\textwidth}
        \centering
        \vspace{2em}
        \includegraphics[width=\linewidth]{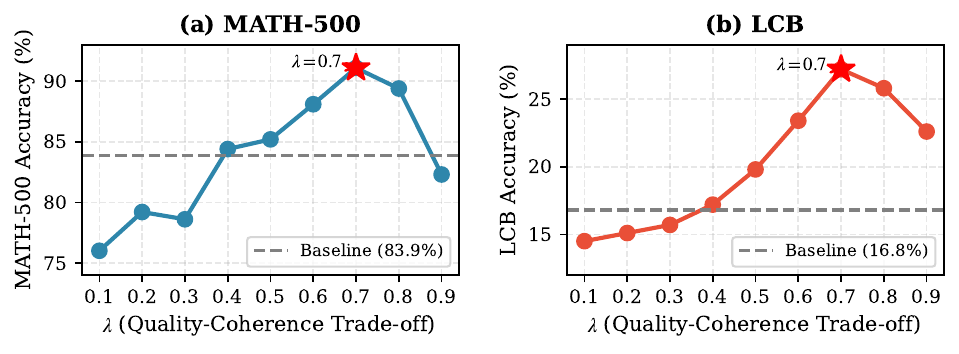}
        \vspace{0em}
        \caption{Sensitivity analysis of $\lambda$ (quality-coherence trade-off). Performance peaks at $\lambda=0.7$; Smaller $\lambda$ values over-prioritize coherence, while larger values produce less coherent chains.}
        \label{fig:lambda_sensitivity}
    \end{minipage}
\end{figure}

\paragraph{Single-Teacher vs Multi-Teacher.}
Figure~\ref{fig:radar_single_multi} illustrates the performance gap between single-teacher distillation and CRD's multi-teacher approach. Both Gemini-Flash-Distill and DeepSeek-R1-Distill apply the same scoring and student-training pipeline to a single teacher's outputs, but remove cross-feedback and cross-teacher step stitching.
CRD consistently outperforms both across all benchmarks, with particularly large gains on AIME'24 (+14--18\%) and LCB (+10\%), showing that combining two heterogeneous teachers outperforms either teacher alone.

\paragraph{Lambda sensitivity.} Figure~\ref{fig:lambda_sensitivity} shows the sensitivity of the coherence-quality trade-off parameter $\lambda$ in Thinking Path.
At $\lambda \leq 0.5$, excessive weight on transition coherence leads to selecting plausible-sounding but low-quality steps (MATH-500 drops to 76.0\%).
At $\lambda > 0.8$, local quality dominates and coherence is underweighted, producing disconnected reasoning chains (82.3\%). The optimal value, $\lambda=0.7$, balances high-quality step selection while maintaining logical flow.
\section{Related Work}
\label{sec:discussion}

\textbf{Reasoning and Distillation.} Transferring reasoning capabilities from large models to smaller students via knowledge distillation is a key research area. Prior work has explored distilling Chain-of-Thought demonstrations~\citep{wei2023chainofthoughtpromptingelicitsreasoning}, leveraging multiple models for enhanced knowledge transfer~\citep{tian2024answerstransferringreasoningcapabilities}, and iterative refinement techniques~\citep{saunders2022selfcritiquingmodelsassistinghuman, liang2024encouragingdivergentthinkinglarge}. Contrastive Representation Distillation~\citep{tian2022contrastiverepresentationdistillation} aligns student and teacher embeddings for finer-grained transfer. Recent work demonstrates that budget forcing techniques can yield significant reasoning improvements with small curated datasets~\citep{muennighoff2025s1simpletesttimescaling}. LIMO~\citep{ye2025limoreasoning} shows that a few hundred carefully curated examples can elicit strong mathematical reasoning, while CRD improves each trajectory through cross-teacher feedback and step-level synthesis across mathematics, coding, logic, and STEM. Building upon these foundations, our Collaborative Reasoning Distillation (CRD) framework generates high-quality reasoning datasets through integrated multi-teacher generation, cross-feedback refinement, and fine-grained quality assessment using multi-judge evaluation.

\textbf{Multi-Teacher Distillation Methods.} Recent work has explored various approaches to leverage multiple teacher models for knowledge distillation. TinyLLM uses multi-task SFT where the student imitates both answers and teacher-specific rationales. FAIR uses peer review among teachers to filter rationales with an acceptance threshold. MAGDi distills multi-agent interaction graphs with a contrastive objective between correct and incorrect reasoning. Our CRD framework takes a complementary approach by introducing interactive refinement (cross-feedback) before distillation and explicit step-wise quality scoring during training. This enables the student to learn from the collaborative synthesis of multiple perspectives, resulting in 3.6--23\% improvements over these methods when using identical training data (see Section~\ref{sec:multi_teacher_comparison}).

\textbf{Efficient Reasoning Distillation.} Advanced reasoning models such as DeepSeek-R1~\citep{Guo_2025} have demonstrated that extended test-time computation can substantially enhance reasoning capabilities.
CRD is complementary to test-time scaling: its primary goal is to improve the quality of distilled supervision, while the Optimal Budget curriculum can additionally encourage shorter student reasoning at inference time.
This aligns with recent findings that concise reasoning can achieve comparable accuracy to verbose chains~\citep{xu2025chaindraftthinkingfaster}, and that reasoning structure often matters more than output length~\citep{he2025valleycodereasoningscaling}. Our Optimal Budget curriculum enables students to produce effective solutions using a fraction of the tokens required by teacher models, making advanced reasoning accessible for latency-sensitive deployments without sacrificing problem-solving capability.

\textbf{LLM-as-a-Judge.} LLM-as-a-Judge serves as a scalable alternative to human evaluation for assessing intermediate reasoning steps~\citep{gu2025surveyllmasajudge}. Unlike outcome-based evaluation, step-level assessment provides dense quality signals that help models learn the distinction between logically sound reasoning and lucky guesses.  EvalPlanner~\citep{saha2025learningplanreason} demonstrates that learned evaluation criteria can effectively assess reasoning quality, providing a foundation for our multi-judge approach.

\section{Conclusion}
\label{sec:conclusion}

This paper presents Collaborative Reasoning Distillation (CRD), a framework that transfers reasoning capabilities from multiple foundation models to compact students through multi-teacher cross-feedback, fine-grained quality assessment, and coherence-aware step synthesis.
CRD achieves strong reasoning performance in small models on the evaluated benchmarks: CRD-4B reaches 97.3\% on MATH-500, 70.3\% on AIME'25, and 57.4\% on LiveCodeBench, using only 50K training examples, 12$\times$ fewer final training trajectories than standard distillation.
CRD demonstrates that step-level synthesis enables solving problems where individual teachers fail.

\section*{Limitations and Future Work}
CRD relies on LLM teachers and judges and may inherit some of their biases, so plausible but incorrect steps can occasionally receive high scores. Stitching selects paths that score well under our quality and coherence criteria, which does not by itself guarantee that every selected chain is logically correct. We evaluate one teacher pair (DeepSeek-R1 and Gemini 2.5 Flash), and other teacher combinations remain to be explored. The reported training cost covers student optimization only and does not include teacher generation or curation. We compare RQO with naive SFT on the same data but not with DPO.

While our current rubric provides a robust baseline, reliance on static criteria limits scalability across diverse domains. A promising future direction is to transition from hand-crafted rules to \textit{learned quality models} trained on preference data. Such models could dynamically adapt to domain-specific nuances and enable more scalable supervision without manual rubric engineering. Furthermore, extending our framework to handle heterogeneous teacher feedback could mitigate potential sycophancy, paving the way for more robust collaborative reasoning systems. Future work includes larger and all-open-weight teacher pools such as QwQ-32B and GPT-OSS, and stronger teachers for expert-level benchmarks such as HLE. We also plan controlled comparisons with on-policy distillation and SFT-based baselines, formal bounds on stitching failures, and a larger human audit of stitched chains.

\bibliographystyle{plainnat}
\bibliography{paper}

\clearpage
\newpage
\appendix

\section{Cross-Feedback Review Prompt}
\label{app:review_prompt}

This section provides the complete prompt template used in the cross-feedback refinement step (Section~\ref{sec:cross_feedback}) to instruct teacher models to critically review peer reasoning chains across mathematical, coding, and logical reasoning tasks.

\begin{lstlisting}[basicstyle=\ttfamily\scriptsize, frame=single, breaklines=true]
[SYSTEM]
You are an expert reasoning evaluator specializing in mathematical problem-solving, code generation, and logical deduction. Your task is to critically review a reasoning chain from another model and generate an improved solution.

Be rigorous and critical in your evaluation. Do not accept superficially correct answers--identify subtle logical gaps, unstated assumptions, and potential edge cases that the original solution may have overlooked.

[PROBLEM]
{problem}

[DOMAIN]
{domain}  # One of: MATH, CODE, LOGIC

[PEER REASONING CHAIN]
The following is a reasoning chain from another teacher model. Your task is to review it critically and improve upon it.

--- Other Teacher's Output ---
{other_teacher_reasoning_chain}

[REVIEW INSTRUCTIONS]
Analyze the peer chain thoroughly. Your review must cover both general reasoning quality and domain-specific criteria.

=== General Reasoning Criteria ===
1. Logical Flow:
   - reasoning:coherence - Are transitions between steps justified?
   - reasoning:completeness - Are all necessary steps present?
   - reasoning:assumptions - Are implicit assumptions stated and valid?

2. Error Detection:
   - reasoning:contradiction - Does any step contradict previous ones?
   - reasoning:circular - Is there circular reasoning?
   - reasoning:leap - Are there unjustified logical leaps?

3. Quality Assessment:
   - reasoning:clarity - Is the reasoning clear and followable?
   - reasoning:efficiency - Is the solution unnecessarily verbose?
   - reasoning:novelty - Does it offer creative insights?

=== Domain-Specific Criteria ===

[If MATH]:
   - math:algebraic_validity - Are algebraic manipulations correct?
   - math:arithmetic - Are calculations accurate?
   - math:constraint_satisfaction - Are problem constraints respected?
   - math:case_coverage - Are all cases (including edge cases) handled?
   - math:theorem_applicability - Are theorems applied with valid preconditions?
   - math:unit_consistency - Are units handled correctly (if applicable)?

[If CODE]:
   - code:compilable - Does the code compile/run without syntax errors?
   - code:correctness - Does it produce correct output for given inputs?
   - code:edge_cases - Are boundary conditions handled?
   - code:efficiency - Is time/space complexity reasonable?
   - code:readability - Is the code well-structured and readable?
   - code:error_handling - Are potential runtime errors handled?

[If LOGIC]:
   - logic:premise_validity - Are premises clearly stated and valid?
   - logic:inference_rules - Are deduction rules correctly applied?
   - logic:quantifier_scope - Are quantifiers (all, some, none) used correctly?
   - logic:negation - Are negations handled properly?
   - logic:counterexample - Has the solution considered counterexamples?
   - logic:soundness - Is the argument logically sound?

[TASK]
Based on your critical review:
1. Identify ALL errors, gaps, and weaknesses (be thorough, not lenient)
2. Note any strengths worth preserving
3. Generate an improved reasoning chain that fixes identified issues

[OUTPUT FORMAT]
<review>
[Critical analysis of the peer chain]
- Strengths: [What to preserve]
- Weaknesses: [Errors, gaps, issues found - be specific with tags]
- Verdict: [ACCEPT_WITH_MINOR_FIXES | NEEDS_MAJOR_REVISION | REJECT_AND_REDO]
</review>

<think>
[Your improved step-by-step reasoning that addresses all identified issues]
</think>

<answer>
[Final answer]
</answer>
\end{lstlisting}

\section{Mathematical Scoring Rubric (T-LCM)}
\label{app:scoring_rubric}

This section provides the full scoring rubric used for mathematical reasoning evaluation. The \textbf{T-LCM (Taxonomy-Guided Logic-Consistency Metric)} protocol operates in two phases: domain classification at the problem level, followed by step-wise scoring with explicit decision boundaries.

\subsection{Score Range and Decision Boundaries}

\begin{table}[ht]
\centering
\small
\caption{T-LCM scoring rubric with decision boundaries for mathematical reasoning.}
\label{tab:tlcm_rubric}
\begin{tabularx}{\textwidth}{@{} c l X l @{}}
\toprule
\textbf{Score} & \textbf{Grade} & \textbf{Description} & \textbf{Decision Boundary} \\
\midrule
10.0 & Rigorous & Flawless execution, all constraints satisfied, all cases covered. & N/A (Perfect) \\
8.0--9.9 & Valid & Correct logic with minor inefficiency (redundancy, non-standard notation). & Does inefficiency obscure argument? \\
6.0--7.9 & Minor Slip & Recoverable execution errors (calculation, sign, transcription, off-by-one). & Can error be fixed by single value change? \\
3.0--5.9 & Structural Flaw & Significant reasoning errors (constraint violation, theorem misuse, missing cases). & Is there any mathematical basis? \\
0.0--2.9 & Critical Failure & Fundamental logic breakdown (hallucination, logical leap, circular reasoning). & Is reasoning entirely baseless? \\
\bottomrule
\end{tabularx}
\end{table}

\subsection{Error Taxonomy}

We categorize errors into three types based on severity and recoverability:

\textbf{Type I (Execution Errors, 6.0--7.9):} Calculation errors where correct formula is applied but arithmetic is wrong; sign errors with correct equation structure; transcription errors copying values incorrectly; off-by-one boundary errors in discrete domains; unit oversight with approximately correct magnitude.

\textbf{Type II (Reasoning Errors, 3.0--5.9):} Constraint violations ignoring core problem conditions; conceptual misapplication using theorems where preconditions are not met; incomplete case analysis missing critical branches; variable scope errors reusing symbols for different meanings; premature approximation causing significant deviation.

\textbf{Type III (Logic Breakdown, 0.0--2.9):} Hallucination inventing non-existent theorems or values; logical leaps concluding without justification; circular reasoning using conclusions as premises; contradiction with previous steps; irrelevance to the problem.

\subsection{Domain-Specific Checklists}

The scoring protocol is domain-adaptive. For \textbf{algebra}, we check sign consistency, factoring validity, inequality direction, and domain restrictions. For \textbf{geometry}, we verify diagram consistency, angle/length properties, and similarity/congruence conditions. For \textbf{number theory}, we validate modular arithmetic, divisibility claims, and integer constraints. For \textbf{combinatorics}, we examine counting methods, overcounting adjustments, and boundary conditions.

\subsection{Full Judge Prompt Template}

Below is the complete prompt template used for mathematical reasoning evaluation, inspired by xVerify~\citep{chen2025xverifyefficientanswerverifier}:

\begin{lstlisting}[basicstyle=\ttfamily\scriptsize, frame=single, breaklines=true]
[SYSTEM]
You are a Mathematics Olympiad Judge. Evaluate the given reasoning step using the T-LCM protocol.

[DOMAIN] {domain}

[RUBRIC]
- 10.0 (Rigorous): Flawless deduction, all constraints satisfied, all cases covered.
- 8.0-9.9 (Valid): Correct logic but contains redundancy or non-standard notation.
- 6.0-7.9 (Minor Slip): Execution errors (calculation, sign, transcription, off-by-one) with correct structure.
- 3.0-5.9 (Structural Flaw): Reasoning errors (constraint violation, theorem misuse, missing cases).
- 0.0-2.9 (Critical Failure): Hallucination, logical leap, circular reasoning, contradiction.

[DECISION BOUNDARIES]
- 8.0 vs 7.9: Does inefficiency obscure the argument? (Yes -> below 8.0)
- 6.0 vs 5.9: Can error be fixed by changing a single value? (Yes -> 6.0+)
- 3.0 vs 2.9: Is there any mathematical basis? (Yes -> 3.0+)

[DOMAIN-SPECIFIC CHECKLIST: {domain}]
{domain_checklist}

[INPUT]
Problem: {problem}
Previous Steps: {history}
Current Step: {current_step}

[OUTPUT FORMAT]
Score: [0.0-10.0]
Error Type: [None | Calculation | Sign | Transcription | Off-by-One | Constraint Violation | Theorem Misuse | Missing Case | Hallucination | Logical Leap | Circular | Contradiction | Irrelevance]
Rationale: [Brief explanation]
\end{lstlisting}

\section{Coherence-Aware Thinking Path Algorithm}
\label{app:thinking_path_algorithm}

This section provides the detailed algorithm for Thinking Path, our coherence-aware pruning method that curates high-quality reasoning steps from multiple teacher outputs while preventing the Frankenstein effect, the selection of high-scoring but logically disconnected steps that form an incoherent chain.

\subsection{Algorithm Overview}

Unlike naive greedy pruning that independently selects the highest-scoring step at each position, our \textbf{Coherence-Aware Thinking Path} algorithm optimizes a joint objective of \textbf{Quality} (from the aggregated JudgeLM and xVerify step scores) and \textbf{Coherence} (transition coherence between consecutive steps). This formulation is solved efficiently using Viterbi-style dynamic programming.

\begin{itemize}
    \item \textbf{Node Score} $q(s)$: The aggregated step-quality score $v_t$ (25\% trimmed mean of the four JudgeLM and xVerify scores; Section~\ref{sec:quality_assessment}), normalized to $[0,1]$.
    \item \textbf{Edge Weight}: The transition-coherence score between consecutive steps $s_{t-1}$ and $s_t$, computed from the cosine similarity of their embeddings and rescaled to $[0,1]$.
\end{itemize}

\textbf{Complexity Analysis.} The time complexity of our Coherence-Aware Path Finding is $O(T \cdot N^2)$, where $T$ is the maximum number of reasoning steps per chain and $N$ is the number of teacher models. Since $N$ is small ($N=2$ in our experiments) and reasoning steps are bounded ($T < 20$ for most problems), the computational overhead is negligible compared to the LLM generation cost of initial generation and cross-feedback. The space complexity is $O(T \cdot N)$ for storing the dynamic programming table and backpointers.

\subsection{Implementation}

\begin{lstlisting}[basicstyle=\ttfamily\scriptsize, frame=single, breaklines=true]
def coherence_aware_thinking_path(chains, lambda_=0.7, tau=7.0):
    chains = align_steps(chains)  # embedding-based step alignment with lookahead merging

    num_steps = max(len(c) for c in chains)
    num_teachers = len(chains)

    # DP Table: dp[t][i] = max cumulative score ending at teacher i, step t
    dp = [[-float('inf')] * num_teachers for _ in range(num_steps)]
    parent = [[-1] * num_teachers for _ in range(num_steps)]

    # Initialization (Step 0)
    for i in range(num_teachers):
        if len(chains[i]) > 0:
            dp[0][i] = normalize(chains[i][0]['score'])  # Scale to 0-1

    # Forward Pass (Viterbi Search)
    for t in range(1, num_steps):
        for curr_i in range(num_teachers):
            if t >= len(chains[curr_i]):
                continue

            # 1. Quality score (aggregated JudgeLM/xVerify step score)
            q_score = normalize(chains[curr_i][t]['score'])

            best_prev_val = -float('inf')
            best_prev_idx = -1

            for prev_i in range(num_teachers):
                if t-1 >= len(chains[prev_i]):
                    continue

                # 2. Coherence score C(u, v) (transition coherence)
                c_score = get_transition_coherence(
                    chains[prev_i][t-1]['content'],
                    chains[curr_i][t]['content']
                )

                # 3. Joint Score: Quality + Coherence
                transition_score = dp[t-1][prev_i] + \
                    (lambda_ * q_score) + \
                    ((1 - lambda_) * math.log(c_score + 1e-9))

                if transition_score > best_prev_val:
                    best_prev_val = transition_score
                    best_prev_idx = prev_i

            dp[t][curr_i] = best_prev_val
            parent[t][curr_i] = best_prev_idx

    # Backward Pass (Path Reconstruction)
    final_idx = num_steps - 1
    best_end = max(range(num_teachers), key=lambda i: dp[final_idx][i])

    optimal_path = []
    curr = best_end
    for t in range(final_idx, -1, -1):
        step = chains[curr][t]
        if step['score'] >= tau:  # Final quality threshold
            optimal_path.append(step)
        curr = parent[t][curr]

    return optimal_path[::-1]  # Reverse to chronological order

def get_transition_coherence(prev_content, curr_content):
    """Compute C(prev, curr) with rescaled embedding cosine similarity."""
    return (1 + cosine_similarity(embed(prev_content), embed(curr_content))) / 2  # rescale [-1, 1] to [0, 1]
\end{lstlisting}

\subsection{Key Design Choices}

The algorithm embodies several important design decisions that distinguish it from simple greedy selection.

The \textbf{joint optimization} of quality and coherence prevents the Frankenstein effect where high-scoring but disconnected steps form an invalid reasoning chain. The hyperparameter $\lambda$ controls this trade-off: higher values ($\lambda > 0.5$) prioritize step quality scores, while lower values emphasize smooth transitions between steps. In our experiments, $\lambda = 0.7$ achieved the best balance, allocating 70\% weight to quality and 30\% to coherence.

The \textbf{Viterbi-style dynamic programming} ensures global optimality in $O(T \cdot N^2)$ time, where $T$ is the maximum chain length and $N$ is the number of teachers. This is efficient for typical settings ($N=2$, $T < 20$).

\section{Training Data Composition}
\label{app:dataset}

Table~\ref{tab:dataset_stats} lists the number of training instances drawn from each source.

\begin{table}[h!]
\caption{Composition of the CRD-50K training dataset.}
\centering
\footnotesize
\begin{tabular}{@{}llc@{}}
\toprule
\textbf{Source Dataset} & \textbf{Domain} & \textbf{Samples}  \\
\midrule
DeepScaleR & Math & 20K \\
Eurus-2-RL & Code  & 15K \\
Reasoning Gym & Logical & 10K  \\
SCP-116K & STEM & 5K  \\
\midrule
\textbf{Total} &  & \textbf{50K}  \\
\bottomrule
\end{tabular}
\vspace{5pt}
\label{tab:dataset_stats}
\end{table}

\section{Benchmark Descriptions}
\label{app:benchmarks}

We evaluate CRD on nine benchmarks spanning mathematics, code generation, STEM reasoning, planning, and expert-level knowledge:

\begin{itemize}
    \item \textbf{MATH-500}~\citep{lightman2023letsverifystepstep}: Competition mathematics problems requiring multi-step reasoning across the seven MATH subjects (prealgebra, algebra, number theory, counting and probability, geometry, intermediate algebra, and precalculus).
    \item \textbf{LiveCodeBench}~\citep{jain2024livecodebenchholisticcontaminationfree}: Code generation benchmark with execution-based evaluation measuring functional correctness.
    \item \textbf{GPQA-Diamond}~\citep{rein2023gpqagraduatelevelgoogleproofqa}: Graduate-level questions in physics, biology, and chemistry requiring deep domain knowledge.
    \item \textbf{AIME'24}~\citep{aime2024} \textbf{and AIME'25}~\citep{aime2025}: American Invitational Mathematics Examination problems spanning algebra, geometry, number theory, and combinatorics that demand creative multi-step solutions.
    \item \textbf{GSM8K}~\citep{cobbe2021trainingverifierssolvemath}: Grade-school math word problems, used for the multi-teacher comparison (Table~\ref{tab:multi_teacher_qwen}).
    \item \textbf{HumanEval}~\citep{chen2021evaluatinglargelanguagemodels}: Python function synthesis evaluated with unit tests, used for the multi-teacher comparison (Table~\ref{tab:multi_teacher_qwen}).
    \item \textbf{BoxNet}~\citep{stojanovski2025reasoninggymreasoningenvironments}: Procedurally generated multi-step planning that moves colored boxes to their targets, used as an out-of-distribution test.
    \item \textbf{HLE}~\citep{phan2026benchmark}: Expert-level questions across many academic fields. We use the text-only subset.
\end{itemize}

\section{Detailed Analysis}
\label{app:detailed_analysis}

\subsection{MATH-500 Subject Breakdown}

MATH-500 covers seven subjects; we analyze algebra, precalculus, and number theory. To quantify the impact of collaborative feedback, we measured per-subject Chain-of-Thought accuracy by matching generated reasoning steps against official solution templates. Algebra shows the clearest improvement: on a 50-problem slice, the DeepSeek-R1-1.5B baseline attains 85.8\% step correctness (482/562 steps), whereas CRD-1.5B reaches 92.2\% (688/746 steps). Precalculus benefits by 6--8 percentage points in function analysis and limit computations, while number theory improves by 5--7 points in divisibility and modular-arithmetic reasoning.

\subsection{Scaling Analysis}

Our results reveal consistent scaling patterns. On the DeepSeek-R1-Distill-Qwen-1.5B student, CRD improves MATH-500 by 7.2 points (83.9 $\rightarrow$ 91.1). For Qwen3 students, MATH-500 gains shrink as the base model saturates (1.7B: +1.1, 4B: +0.3, 30B-A3B: +0.3), while gains on the harder AIME'25 persist (+5.4, +4.7, +2.3), suggesting that the marginal benefit of CRD diminishes on near-saturated benchmarks but remains on harder ones.

\section{Multi-Round Cross-Feedback Analysis}
\label{sec:appendix_multi_round}

We analyze the effect of multiple cross-feedback rounds on final model performance.

\begin{table}[ht]
\centering
\small
\caption{Impact of cross-feedback rounds on model performance. Evaluated on Qwen3-1.7B student model. Performance converges at round 2--3.}
\label{tab:multi_round}
\begin{tabular}{@{}lcccc@{}}
\toprule
\textbf{Round} & \textbf{MATH-500} & \textbf{GPQA-Diamond} & \textbf{LCB} & \textbf{Marginal Gain} \\
\midrule
Initial (0) & 93.4 & 40.1 & 33.2 & -- \\
Round 1 & 93.8 & 41.5 & 34.7 & +0.4 / +1.4 / +1.5 \\
Round 2 & 94.0 & 42.2 & 35.1 & +0.2 / +0.7 / +0.4 \\
Round 3 & 94.0 & 42.2 & 35.0 & 0 / 0 / -0.1 \\
\bottomrule
\end{tabular}
\end{table}

Performance converges at round 2--3, with additional rounds providing no benefit. The marginal utility of cross-feedback diminishes rapidly, suggesting that 2 rounds achieve the optimal quality-efficiency trade-off.

\section{Judge Reliability}
\label{app:judge}
We reduce dependence on any single evaluator with two heterogeneous judges (JudgeLM and xVerify), two evaluations per judge, a 25\% trimmed mean over the four scores, and domain-specific verification (execution checks for code and answer-equivalence checks for mathematics). Re-scoring all steps in our evaluation set with GPT-4o, given the same preceding history and current step, yields a mean absolute difference of 0.428 from our aggregated score on the 0--10 scale. On MATH-500, the full dual-judge configuration improves the student by 1.4 points, compared with 0.6 with xVerify alone, 0.4 with JudgeLM alone, and 0.6 with GPT-4o alone. Correlated evaluator biases may remain, and plausible but incorrect reasoning may still receive high scores.

\section{Stitched Training Example}
\label{app:stitching}
Table~\ref{tab:stitch_example} shows a stitched training trajectory for a problem that asks for the fraction of an equilateral triangle lying outside a circle tangent to two of its sides. The original chain derives the first two steps correctly but then assumes only an exterior-center configuration. Cross-feedback produces a corrected continuation, and Thinking Path selects $A_1 \rightarrow A_2 \rightarrow B_3 \rightarrow B_4 \rightarrow B_5$.

\begin{table}[h]
\centering
\caption{A stitched training trajectory ($A$: original chain, $B$: cross-feedback continuation).}
\label{tab:stitch_example}
\small
\begin{tabular}{@{}llcc@{}}
\toprule
\textbf{Step} & \textbf{Content} & \textbf{Score} & \textbf{Decision} \\
\midrule
$A_1$ & Derives $\angle BOC = 120^\circ$ & 9.3 & Retain \\
$A_2$ & Derives $r = s/\sqrt{3}$ & 9.0 & Retain \\
$A_3$ & Assumes only an exterior center & 3.6 & Reject \\
$A_4$ & Applies an absolute value to repair a negative area & 0.8 & Reject \\
$B_3$ & Identifies two feasible center configurations & 9.4 & Retain \\
$B_4$ & Rejects the exterior case by contradiction & 9.1 & Retain \\
$B_5$ & Computes the valid minor-segment area & 8.9 & Retain \\
\bottomrule
\end{tabular}
\end{table}

\section{Student Output Examples}
\label{app:student_outputs}
We compare DeepSeek-R1-Distill-Qwen-1.5B and CRD-1.5B on a held-out MATH-500 problem with identical prompts and decoding. The problem asks for $p(8)$, where $p(x)$ is a polynomial of degree 5 such that $p(n) = \frac{n}{n^2-1}$ for $n = 2, 3, \ldots, 7$.

\textbf{DeepSeek-R1-Distill-Qwen-1.5B (incorrect).} The model defines $q(x) = x^2 - 1 - p(x)$ and asserts $q(2) = \cdots = q(7) = 0$, which does not follow from the condition. It then writes this degree-5 polynomial as a product of six linear factors and outputs $p(8) = -657$.

\textbf{CRD-1.5B (correct).} The model constructs $Q(x) = (x^2-1)p(x) - x$, notes that $Q(n) = 0$ for $n = 2, \ldots, 7$, keeps the degree-7 structure, and uses $Q(1) = -1$ and $Q(-1) = 1$ to determine the remaining linear factor. It obtains $Q(8) = -259/56$ and $63p(8) - 8 = -259/56$, and returns $p(8) = 3/56$.

\end{document}